\documentclass[10pt,twocolumn,letterpaper]{article}

\usepackage[margin=0.75in]{geometry}

\usepackage[utf8]{inputenc}
\usepackage[T1]{fontenc}
\usepackage{times}
\usepackage{hyperref}
\usepackage{url}
\usepackage{booktabs}
\usepackage{amsfonts}
\usepackage{amsmath}
\usepackage{abstract}
\usepackage{amssymb}
\usepackage{nicefrac}
\usepackage{microtype}
\usepackage{xcolor}
\usepackage{graphicx}
\usepackage{multirow}
\usepackage{makecell}
\usepackage{algorithm}
\usepackage{algpseudocode}
\usepackage{subcaption}
\usepackage{float}
\usepackage{placeins}
\usepackage{pgfplots}
\usepackage{wrapfig}
\usepackage{enumitem}
\usepackage{pifont}
\usepackage{fontawesome5}
\usepackage{tikz}
\usetikzlibrary{shapes, positioning, arrows.meta, calc, fit, backgrounds}
\usepackage[round]{natbib}
\usepackage{titlesec}
\usepackage[table]{xcolor}
\setlist{nosep}

\titlespacing*{\subsection}{0pt}{3pt}{1pt}
\titlespacing*{\subsubsection}{0pt}{2pt}{0.5pt}
\titlespacing*{\section}{0pt}{6pt}{3pt}

\usepackage{listings} 

\lstdefinestyle{promptstyle}{
    basicstyle=\small\ttfamily, 
    breaklines=true,            
    breakatwhitespace=true,     
    frame=single,               
    rulecolor=\color{black},    
    backgroundcolor=\color{gray!5}, 
    columns=flexible,            
    keepspaces=true,            
    showstringspaces=false,     
    captionpos=b                
}

\titleformat{\section}{\large\bfseries}{\thesection}{1em}{}
\titleformat{\subsection}{\normalsize\bfseries}{\thesubsection}{1em}{}
\titleformat{\subsubsection}{\normalsize\itshape}{\thesubsubsection}{1em}{}
\hypersetup{
    colorlinks=true,
    linkcolor=blue!70!black,
    citecolor=green!50!black,
    urlcolor=blue!70!black
}

\definecolor{PanguColor}{RGB}{70, 130, 180}
\definecolor{QwenColor}{RGB}{220, 88, 88}
\definecolor{WinColor}{RGB}{34, 139, 34}
\definecolor{LightGray}{RGB}{245, 245, 245}

\newcommand{\pangu}{\textsc{OpenPangu-7B}}
\newcommand{\qwen}{\textsc{Qwen2.5-7B-Instruct}}
\newcommand{\ssc}{\textsc{SSC}}

\usetikzlibrary{patterns}
\pgfplotsset{compat=1.17}

\title{\textbf{Structured Self-Consistency:}\\
A Multi-Task Evaluation of LLMs on VirtualHome}

\author{Jiaqi Xu\textsuperscript{1}, Tao Huang\textsuperscript{2},Kai Zhang\textsuperscript{$\dagger$}\\
\faGithub\ \href{https://github.com/HsuJQ/SSC}{https://github.com/HsuJQ/SSC}}

\date{}

\begin{document}

\maketitle

\begingroup
  \def\thefootnote{$\dagger$}\footnotetext{Corresponding author}
\endgroup

\begin{abstract}
\noindent
Embodied AI requires agents to understand goals, plan actions, and execute tasks in simulated environments. We present a comprehensive evaluation of Large Language Models (LLMs) on the VirtualHome benchmark using the Embodied Agent Interface (EAI) framework. We compare two representative 7B-parameter models — \pangu{} and \qwen{} — across four fundamental tasks: \textit{Goal Interpretation}, \textit{Action Sequencing}, \textit{Subgoal Decomposition}, and \textit{Transition Modeling}. We propose \textbf{Structured Self-Consistency (\ssc{})}, an enhanced decoding strategy that leverages multiple sampling with domain-specific voting mechanisms to improve output quality for structured generation tasks. Experimental results demonstrate that \ssc{} significantly enhances performance, with \pangu{} excelling at hierarchical planning while \qwen{} show advantages in action-level tasks. Our analysis reveals complementary strengths across model types, providing insights for future embodied AI system development.
\end{abstract}

\FloatBarrier

\section{Introduction}

Embodied Artificial Intelligence (Embodied AI) represents a frontier of AI research in which agents are required to perceive, reason, and act within physical or simulated environments~\citep{duan2022survey}. In contrast to traditional AI systems that operate primarily on abstract or symbolic data, embodied agents must ground language understanding in physical interactions, placing stringent demands on planning, state tracking, and action execution.

\begin{figure*}[t]
    \centering
    \includegraphics[width=0.9\textwidth]{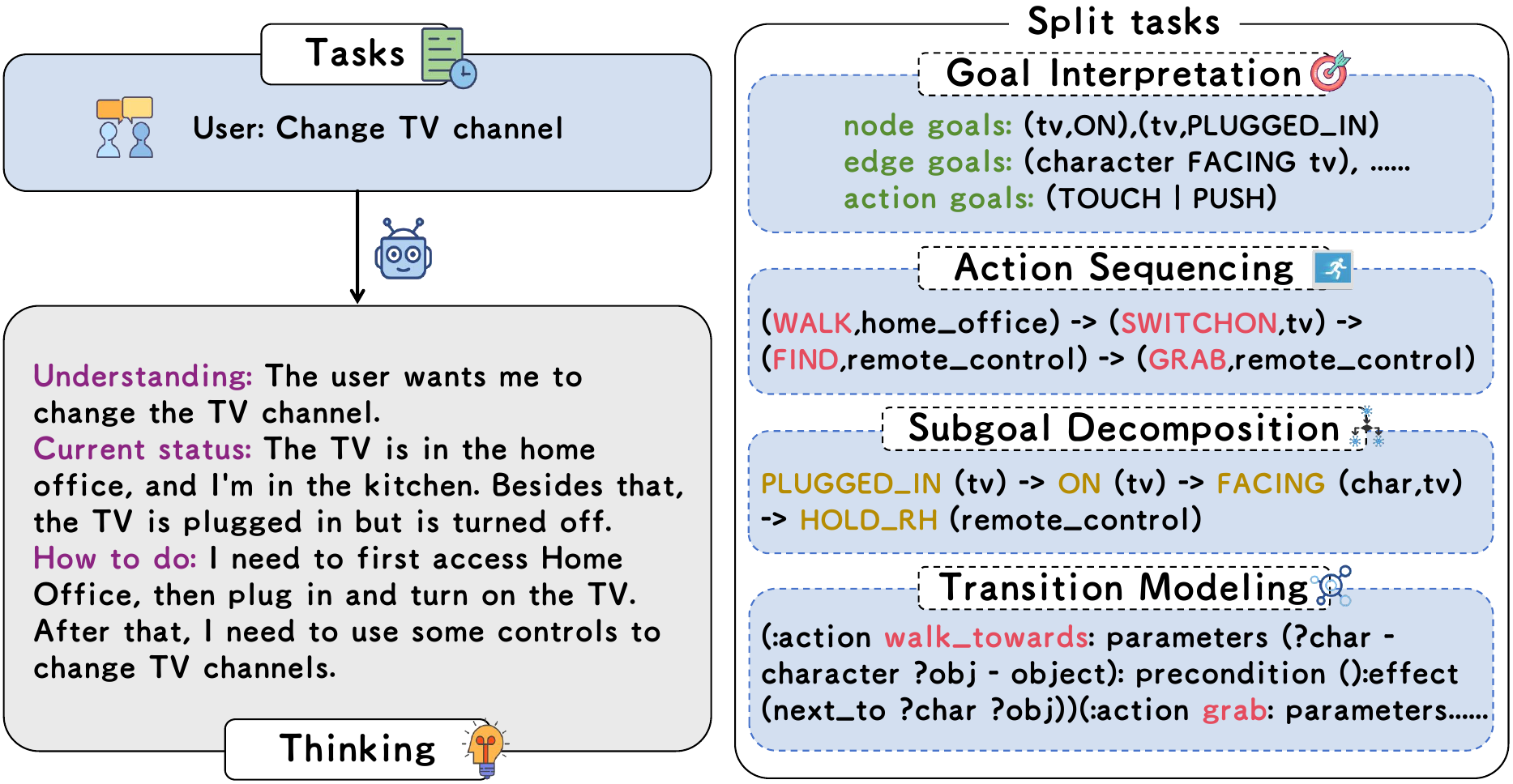}
    \caption{Overview of the four tasks in VirtualHome. Each task targets a distinct aspect of embodied intelligence: Goal Interpretation extracts structured goals from natural language instructions, Action Sequencing generates executable action programs, Subgoal Decomposition decomposes complex tasks into intermediate states, and Transition Modeling predicts action preconditions and effects.}
    \label{fig:image1}
\end{figure*}

The VirtualHome environment~\citep{puig2018virtualhome} provides a realistic household simulation platform for evaluating embodied agents through long-horizon, goal-directed activities. When combined with the Embodied Agent Interface (EAI)~\citep{li2024embodied}, it enables a standardized and modular evaluation framework that captures multiple dimensions of embodied intelligence, as illustrated in Figure~\ref{fig:image1}.

Recent advances in Large Language Models (LLMs) have demonstrated strong capabilities in reasoning and task planning~\citep{mann2020language, ahn2022can}. However, applying LLMs to embodied AI introduces a set of challenges that fundamentally distinguish embodied decision-making from conventional natural language processing tasks:

\begin{itemize}[leftmargin=*, itemsep=4pt, topsep=4pt]
    \item \textbf{Structured Output Generation}: Embodied tasks require outputs to follow strict, domain-specific schemas rather than free-form text. Action sequences must conform to predefined action vocabularies (e.g., \texttt{[WALK]}), while transition models must satisfy valid PDDL syntax. Even minor formatting errors—such as missing brackets or invalid object references—can lead to complete execution failures in the simulator.

    \item \textbf{Physical Grounding and Commonsense Reasoning}: Models must align abstract linguistic instructions with physical constraints, including object affordances (e.g., a \texttt{fridge} can be opened, whereas a \texttt{table} cannot) and spatial relations. Critically, implicit preconditions that humans take for granted—such as opening a container before retrieving an object—must be made explicit for successful embodied execution.

    \item \textbf{Multi-step Planning and Error Propagation}: Household activities typically require long-horizon plans consisting of 10--50 atomic actions. Errors made early in the sequence, such as failing to navigate before interacting with an object, can propagate and invalidate the entire plan. Effective embodied planning therefore requires consistent state tracking and reasoning over subgoal dependencies across extended action horizons.

    \item \textbf{Ambiguity Resolution}: Natural language instructions are often underspecified. For instance, an instruction like ``make breakfast'' may correspond to multiple plausible action sequences depending on contextual assumptions. Models must resolve such ambiguities by inferring reasonable defaults while remaining adaptable to environmental constraints encountered during execution.
\end{itemize}

In this work, we present a systematic evaluation of two representative LLMs, \pangu{}\citep{chen2025pangu} and \qwen{}\citep{qwen2024qwen25}, on the VirtualHome benchmark under the EAI framework. To address the challenges of structured generation in embodied settings, we introduce \textbf{Structured Self-Consistency (\ssc{})}, an inference-time decoding framework that extends the self-consistency paradigm~\citep{wang2022self} with task-specific validation and structure-aware voting mechanisms.

\textbf{Contributions.}
\begin{enumerate}[leftmargin=*, itemsep=3pt, topsep=3pt]
    \item \textbf{Comprehensive Multi-Task Evaluation}: We conduct a unified evaluation of \pangu{} and \qwen{} across four fundamental embodied tasks—Goal Interpretation, Action Sequencing, Subgoal Decomposition, and Transition Modeling—using 338 test instances per task.

    \item \textbf{Structured Self-Consistency (\ssc{})}: We propose a decoding strategy that integrates domain-specific validation with skeleton-based voting for structured outputs, yielding improvements of up to +7.3\% in task success rate and +12.1\% in execution success rate over greedy decoding.

    \item \textbf{Empirical Insights}: Our analysis reveals complementary strengths across models: \pangu{} demonstrates superior hierarchical planning with substantially lower hallucination rates (2.3--5.9\%), while \qwen{} excels in format compliance and action-level execution, achieving up to +8.2\% higher execution success rate.
\end{enumerate}

\section{Related Work}

\begin{figure*}[t]
    \centering
    \includegraphics[width=0.9\textwidth]{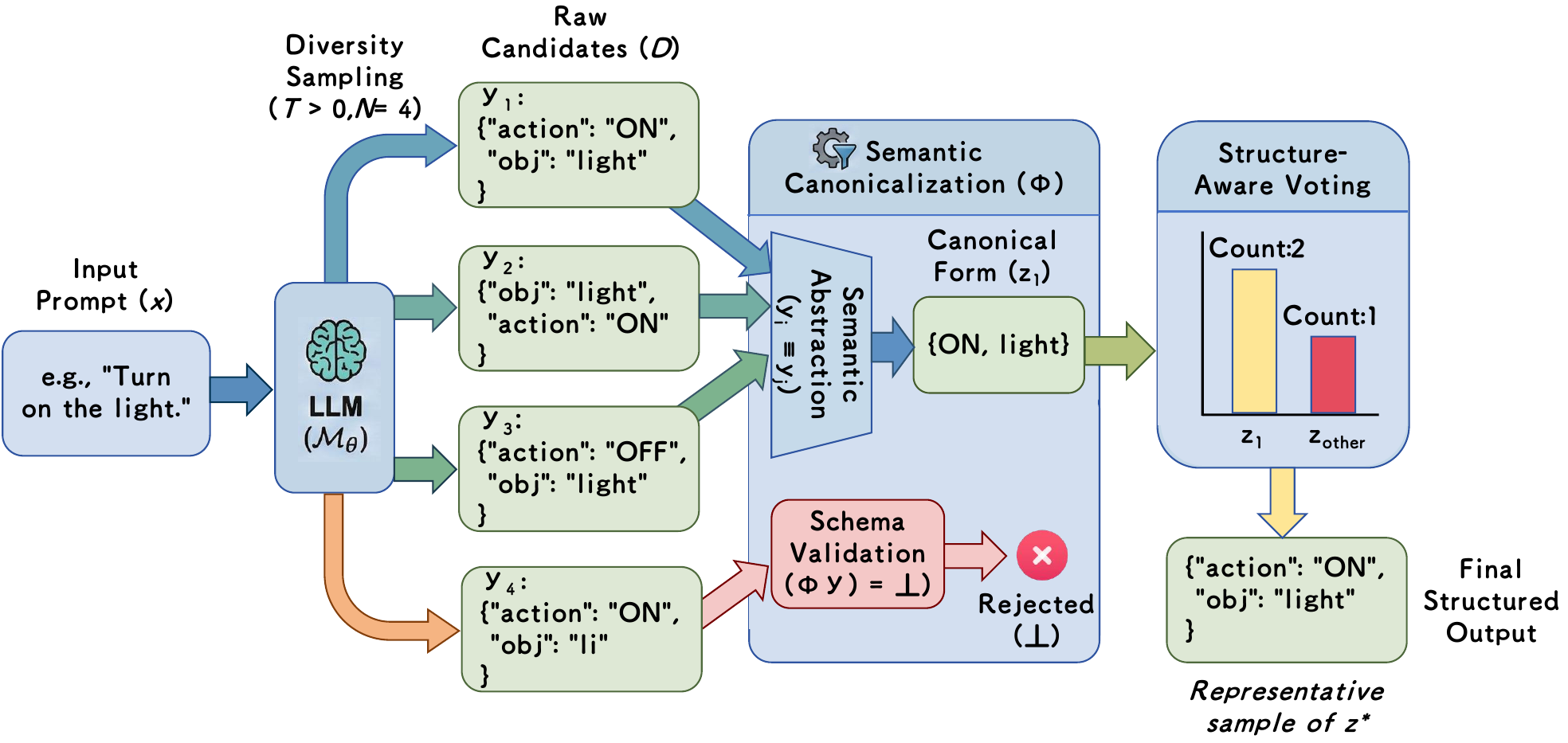}
\caption{Overview of the Structured Self-Consistency (SSC) Methodology, \textbf{illustrated with $N=4$ for simplicity (experiments use $N=5$)}. 
The framework handles two types of noise: 
(1) Syntactic Errors (e.g., broken JSON) are intercepted by the Schema Validator ($\Phi(y)=\bot$). 
(2) Semantic Instability (e.g., $y_3$ generating "OFF" instead of "ON") is resolved via structure-aware voting, where the minority hallucination ($Z_{other}$) is outvoted by the consistent consensus ($Z^*$).}
    \label{fig:image2}
\end{figure*}

\textbf{Embodied AI and Simulation Environments.}
Realistic simulation environments are pivotal for embodied AI. VirtualHome~\citep{puig2018virtualhome} supports complex household activities via programs, while AI2-THOR~\citep{kolve2017ai2} offers physics-enabled 3D environments for navigation and manipulation. Habitat~\citep{savva2019habitat} emphasizes efficiency for large-scale training. Recently, the Embodied Agent Interface (EAI)~\citep{li2024embodied} established a standardized framework unifying task definitions across simulators. Unlike prior work focused on single tasks, we leverage EAI's multi-task protocol to comprehensively assess LLM capabilities across the embodied intelligence spectrum.

\textbf{LLMs for Robotic Task Planning.}
LLMs demonstrate strong potential as robotic planners. SayCan~\citep{ahn2022can} introduced affordance-aware planning, Code as Policies~\citep{liang2022code} enabled control code generation, ProgPrompt~\citep{singh2022progprompt} produced executable programs, and Inner Monologue~\citep{huang2022language} incorporated interactive feedback. While most studies~\citep{ahn2022can, liang2022code, singh2022progprompt, huang2022language} focus on proprietary models or end-to-end execution metrics. In contrast, we systematically compare open-source LLMs across diverse embodied tasks to reveal complementary strengths.

\textbf{Self-Consistency and Decoding Strategies.}
Self-Consistency~\citep{wang2022self} improved reasoning via sampling and majority voting. Extensions include Universal Self-Consistency~\citep{chen2023universal} for free-form outputs, Self-Verification~\citep{weng2023large} for checking, and execution-based selection~\citep{chen2021evaluating} for code correctness. Building on these, our Structured Self-Consistency (\ssc{}) introduces domain-specific validation and skeleton-based voting to address the unique challenges of structured generation in embodied AI.

\section{Methodology}
\label{sec:Methodology}

In this section, we present the theoretical framework of our approach. We first formalize the structured generation problem and then introduce \textit{Structured Self-Consistency} (\ssc{}), a decoding algorithm designed to ensure semantic robustness in constrained output spaces.

\subsection{Problem Formulation}

Let $\mathcal{X}$ denote the domain of natural language instructions and $\mathcal{Y}$ denote the space of valid structured outputs. The validity of any $y \in \mathcal{Y}$ is governed by a deterministic schema $\mathcal{S}$ (e.g., JSON syntax, PDDL predicates). 

We consider a probabilistic language model $\mathcal{M}_\theta$ that models the conditional probability $P(y|x)$. In standard generation, the objective is to find the sequence $\hat{y}$ that maximizes the likelihood:
\begin{align}
    \hat{y} = \operatorname*{arg\,max}_{y \in \mathcal{Y}} P(y|x)
\end{align}
However, in structured reasoning tasks, multiple syntactically distinct sequences can represent the exact same semantic intent (e.g., order-invariant key-value pairs). We denote the semantic equivalence relation as $y_i \equiv y_j$. Our objective is effectively to maximize the likelihood over the equivalence class $[y]_\equiv$, rather than the raw token sequence.

\subsection{Structured Self-Consistency (\ssc{})}
\label{sec:ssc}

Standard greedy decoding often yields suboptimal results in structured tasks due to its sensitivity to surface-level syntactic noise. \ssc{} addresses this by aggregating predictions over semantic clusters. The method consists of three components: \textit{Diversity Sampling}, \textit{Semantic Canonicalization}, and \textit{Structure-Aware Voting}.

\subsubsection{Diversity Sampling}
To approximate the distribution of equivalence classes, we first generate a set of $N$ candidate outputs from the model $\mathcal{M}_\theta$ via stochastic sampling:
\begin{align}
    \mathcal{D} = \{y_1, y_2, \dots, y_N\}, \quad \text{where } y_i \sim P(\cdot|x; \tau)
\end{align}
Here, $\tau > 0$ is the temperature parameter controlling the diversity of the generated hypotheses.

\subsubsection{Semantic Canonicalization}
The core differentiation of \ssc{} lies in the canonicalization function $\Phi: \mathcal{Y} \to \mathcal{Z} \cup \{\bot\}$. This function serves two strictly structured roles that distinguish it from standard self-consistency:

\begin{itemize}[leftmargin=*]
    \item \textbf{Structure Validation:} $\Phi$ acts as a schema gatekeeper. If a generated output $y$ violates the syntax constraints (e.g., malformed JSON, invalid PDDL predicates, or hallucinated actions not in the library), we define $\Phi(y) = \bot$. These invalid pathways are pruned before voting, ensuring the final output is executable.
    
    \item \textbf{Semantic Abstraction:} It maps syntactically distinct but semantically equivalent outputs to a unique signature $z$. For example, in set-based generation, $\Phi(\{a, b\}) = \Phi(\{b, a\})$. This ensures the voting process operates on the \textit{meaning} of the structure rather than its textual representation.
\end{itemize}

\subsubsection{Structure-Aware Voting}
We select the final output by identifying the most probable valid semantic class. The optimal canonical form $z^*$ is obtained by:
\begin{align}
    z^* = \operatorname*{arg\,max}_{z \in \mathcal{Z}, z \neq \bot} \sum_{y_i \in \mathcal{D}} \mathbb{I}[\Phi(y_i) = z]
\end{align}
where $\mathbb{I}[\cdot]$ is the indicator function. The final structured output $\hat{y}$ is selected as the representative sample from $\mathcal{D}$ that corresponds to $z^*$.

The detailed inference procedure, including the voting and filtering logic, is formalized in Algorithm~\ref{alg:ssc} (see Appendix~\ref{app:algorithm}).

\subsection{Task-Adaptive Prompt Engineering}
\label{sec:prompt_engineering}

Standard instruction-following prompts often fail to constrain general-purpose LLMs within the strict bounds of embodied environments. To address this, we designed task-specific system prompts that act as "soft constraints" prior to decoding. Our prompting strategy focuses on two dimensions:

{Domain Logic Injection.} 
For planning tasks like \textit{Action Sequencing} and \textit{Subgoal Decomposition}, models often hallucinate physically impossible actions (e.g., grabbing an object from a distance). We explicitly injected Physical Proximity Rules (e.g., \textit{"To interact with ANY object, the character must be NEXT\_TO that object first"}) and State Consistency Rules into the system prompt. This reduces the search space by pruning logically invalid plans early in the generation process.

\textbf{Syntax Enforcement.} 
Different tasks require distinct output schemas. For \textit{Transition Modeling}, the prompt strictly enforces PDDL syntax (e.g., correct nesting of \texttt{:precondition} and \texttt{:effect}). For \textit{Goal Interpretation} and \textit{Action Sequencing}, we enforce strict JSON schemas, explicitly forbidding conversational fillers or markdown formatting that typically disrupt automated parsers.

The full prompt templates for each task are detailed in Appendix~\ref{app:prompts}.
\section{Evaluations}
We conduct all evaluations on the VirtualHome environment using the Embodied Agent Interface (EAI) framework. All experiments are performed on a unified heterogeneous computing platform to ensure fairness and reproducibility. The system is equipped with an ARM-based HUAWEI Kunpeng 920 (5250) processor, providing 192 CPU cores primarily used for data preprocessing and task scheduling, together with 1.5~TB of system memory.

For accelerated computation, the platform includes eight Ascend 910B2 Neural Processing Units (NPUs), each featuring 64~GB of high-bandwidth memory (HBM), which are utilized for deep learning model training and inference. The operating system is a 64-bit Linux distribution running on the ARM \texttt{aarch64} architecture. All models are implemented in PyTorch (Ascend NPU-adapted version) and accelerated using the Ascend computing platform.

All models are deployed via vLLM with a context window of 4096 tokens. For \ssc{}, we use a sample size of $N=5$ and a temperature of $T=0.7$, together with the task-adaptive canonicalization strategies described in Section~\ref{sec:Methodology}.

\begin{table*}[htbp]
\centering
\caption{\textbf{Main Results on VirtualHome.} We report key metrics for each task. \textbf{Improv.} denotes the relative improvement of SSC over the baseline (Greedy). Note that \pangu{} achieves substantial gains in precision and structured metrics (Transition Modeling F1 +581\%), while \qwen{} benefits significantly in recall-oriented tasks.}
\label{tab:main_results}
\renewcommand{\arraystretch}{1.1}
\resizebox{0.9\textwidth}{!}{
\begin{tabular}{l l | ccc | ccc}
\toprule
\multirow{2}{*}{\textbf{Task}} & \multirow{2}{*}{\textbf{Metric}} & \multicolumn{3}{c|}{\textbf{OpenPangu-7B}} & \multicolumn{3}{c}{\textbf{Qwen2.5-7B}} \\
 & & Baseline & \textbf{+SSC} & \textit{Improv.} & Baseline & \textbf{+SSC} & \textit{Improv.} \\
\midrule
\multirow{3}{*}{\textbf{Goal Interpretation}} 
 & F1 Score(\%) & 21.36 & \textbf{34.29} & \textcolor{red}{+60.5\%} & 24.20 & 26.49 & \textcolor{blue}{+9.5\%} \\
 & Precision(\%) & 30.08 & \textbf{53.94} & \textcolor{red}{+79.3\%} & 49.23 & 17.86 & -63.7\% \\
 & Recall(\%) & 16.56 & \textbf{25.13} & \textcolor{red}{+51.7\%} & 16.05 & 51.25 & \textcolor{blue}{+219.3\%} \\
\midrule
\multirow{2}{*}{\textbf{Action Sequencing}} 
 & Task Success Rate(\%) & 14.43 & \textbf{32.59} & \textcolor{red}{+125.8\%} & 32.46 & 35.74 & \textcolor{blue}{+10.1\%} \\
 & Exec. Success Rate(\%) & 17.70 & \textbf{34.10} & \textcolor{red}{+92.7\%} & 37.40 & 42.30 & \textcolor{blue}{+13.1\%} \\
\midrule
\multirow{2}{*}{\textbf{Subgoal Decomposition}} 
 & Task Success Rate(\%) & 40.63 & \textbf{73.33} & \textcolor{red}{+80.5\%} & 57.99 & 63.02 & \textcolor{blue}{+8.7\%} \\
 & Exec. Success Rate(\%) & 58.50 & \textbf{83.33} & \textcolor{red}{+42.4\%} & 70.41 & 77.81 & \textcolor{blue}{+10.5\%} \\
\midrule
\multirow{3}{*}{\textbf{Transition Modeling}} 
 & F1 Score(\%) & 6.18 & \textbf{42.10} & \textcolor{red}{+581.2\%} & 16.17 & 44.29 & \textcolor{blue}{+173.9\%} \\
 & Precision(\%) & 12.34 & \textbf{56.38} & \textcolor{red}{+356.9\%} & 25.77 & 57.59 & \textcolor{blue}{+123.5\%} \\
 & Recall(\%) & 4.13 & \textbf{30.23} & \textcolor{red}{+632.2\%} & 11.79 & 35.98 & \textcolor{blue}{+205.2\%} \\
\bottomrule
\end{tabular}
}
\end{table*}

\subsection{Experimental Settings}

\textbf{Benchmark.}
We evaluate all models using the Embodied Agent Interface (EAI) benchmark, a unified evaluation framework designed to assess large language models (LLMs) in embodied decision-making scenarios. EAI standardizes task formulations, input-output schemas, and evaluation protocols across embodied environments, enabling fair and systematic comparison of model capabilities without requiring task-specific fine-tuning or environment-dependent adaptations.
In this work, EAI is instantiated on the VirtualHome simulator, which provides a realistic household environment with a fixed action library and explicit object states. The benchmark evaluates LLMs across four fundamental tasks, each targeting a distinct component of embodied intelligence:

\begin{itemize}[leftmargin=*, itemsep=4pt, topsep=4pt]
    \item \textbf{Goal Interpretation (GI).} This task evaluates an agent’s ability to translate free-form natural language instructions into a structured set of goal predicates that define the desired final world state. The output must strictly conform to a predefined schema, and semantic correctness is measured by comparing predicted goal predicates against ground-truth sets in a canonicalized, order-invariant manner.

    \item \textbf{Action Sequencing (AS).} Given an instruction and an initial environment state, the model is required to generate a sequence of low-level executable actions from the VirtualHome action library. This task assesses long-horizon planning, physical grounding, and state tracking, as the generated program must both execute without errors and achieve the specified goal state in the simulator.

    \item \textbf{Subgoal Decomposition (SD).} This task focuses on hierarchical planning. Models must decompose a complex instruction into a sequence of intermediate subgoals, each representing a valid and reachable state that incrementally leads to the final objective. Successful predictions require logical consistency between subgoals as well as executability when translated into action sequences.

    \item \textbf{Transition Modeling (TM).} Transition Modeling evaluates a model’s understanding of environment dynamics by requiring it to predict the preconditions and effects of atomic actions in a PDDL-like formalism. Outputs must adhere to strict syntactic and semantic constraints, making this task particularly sensitive to structural errors and hallucinated predicates.
\end{itemize}

For each task, we use the official EAI test split, consisting of 338 instances per task. Each instance includes a natural-language instruction paired with task-specific structured supervision. All tasks share a common evaluation interface and simulator configuration, ensuring that performance differences reflect model reasoning and generation quality rather than environment-specific heuristics.

\textbf{Baselines.}
We consider two representative open-source large language models with 7B parameters as baselines: \qwen{} and \pangu{}. Both models are evaluated in a zero-shot setting without any task-specific fine-tuning, allowing us to isolate the effect of decoding strategies and structured reasoning frameworks.

\begin{itemize}[leftmargin=*, itemsep=4pt, topsep=4pt]
    \item \textbf{\qwen{}\citep{qwen2024qwen25}}
    is a general-purpose instruction-following model trained on large-scale multilingual and code-rich corpora. Owing to its strong instruction alignment, it demonstrates robust compliance with structured output formats and stable low-level action generation. As a result, \qwen{} typically achieves strong baseline performance on tasks requiring strict schema adherence and executable action sequences, making it a competitive generalist baseline for embodied decision-making.

    \item \textbf{\pangu{}\citep{chen2025pangu}}
    is a general-purpose foundation model with comparatively weaker instruction-following and formatting robustness in zero-shot settings. Its raw outputs often suffer from schema violations or surface-level inconsistencies. Nevertheless, \pangu{} exhibits strong latent semantic understanding and hierarchical planning capabilities, particularly in tasks involving long-horizon reasoning and subgoal structure. This makes it an ideal testbed for examining whether structured inference frameworks can unlock the underlying reasoning potential of less instruction-aligned models.
\end{itemize}

Our proposed method, \ssc{}, is applied on top of both base models as an inference-time framework. Importantly, \ssc{} does not modify model parameters, training data, or prompt templates. All performance gains are achieved purely through structured sampling, validation, and voting during decoding, ensuring a fair and controlled comparison with standard greedy inference.

\begin{table*}[htbp]
\centering
\caption{Fine-grained performance comparison on the Goal Interpretation task (\%).}
\label{tab:goal_interpretation_fine}
\resizebox{0.9\textwidth}{!}{
\begin{tabular}{c ccc ccc ccc}
\toprule
\multirow{2}{*}{\textbf{Model}}
& \multicolumn{3}{c}{\textbf{Node-level}} 
& \multicolumn{3}{c}{\textbf{Edge-level}} 
& \multicolumn{3}{c}{\textbf{Action-level}} \\
\cmidrule(lr){2-4} \cmidrule(lr){5-7} \cmidrule(lr){8-10}
& Precision & Recall & F1 
& Precision & Recall & F1 
& Precision & Recall & F1 \\
\midrule
Qwen2.5-7B 
& 26.72 & 57.00 & 36.38 
& 18.45 & 28.25 & 22.32 
& 9.00  & 72.34 & 16.01 \\

Qwen2.5-7B + SSC
& 24.75 & \textbf{64.41} & 35.76 
& 15.07 & 27.52 & 19.48 
& 12.59 & 67.28 & 21.21 \\

OpenPangu-7B 
& 25.38 & 38.46 & 30.58 
& 16.39 & 13.04 & 14.53 
& 9.72  & 46.67 & 16.09 \\

OpenPangu-7B + SSC
& \textbf{29.23} & 54.63 & \textbf{38.09} 
& \textbf{23.10} & \textbf{28.91} & \textbf{25.68} 
& \textbf{22.51} & \textbf{99.36} & \textbf{36.71} \\
\bottomrule
\end{tabular}
}
\end{table*}

\textbf{Evaluation Metrics.}
We adopt task-specific evaluation metrics defined by the EAI framework to comprehensively assess model performance from three complementary perspectives: \emph{semantic correctness}, \emph{structural validity}, and \emph{executability}. All metrics are computed on the official EAI test split and reported as percentages.

\begin{itemize}[leftmargin=*, itemsep=4pt, topsep=4pt]
    \item \textbf{Goal Interpretation (GI).}
    We evaluate predicted goal predicates using set-based \textbf{Precision}, \textbf{Recall}, and \textbf{F1} score. Predictions and ground-truth goals are first canonicalized to remove order sensitivity and normalize equivalent predicate forms. Precision measures the fraction of predicted predicates that are correct, Recall measures the fraction of ground-truth predicates that are successfully recovered, and F1 is their harmonic mean.

    \item \textbf{Action Sequencing (AS).}
    We report two complementary execution-based metrics:
        \textbf{Task Success Rate (TSR):} the percentage of instances in which the generated action sequence successfully achieves the target goal state in the VirtualHome simulator.
        \textbf{Execution Success Rate (ESR):} the percentage of instances in which the generated program executes to completion without runtime failures, such as invalid actions, missing objects, or violated preconditions.

    While TSR reflects end-task effectiveness, ESR isolates syntactic correctness and physical feasibility, making it particularly sensitive to structural errors in action generation.

    \item \textbf{Subgoal Decomposition (SD).}
    We evaluate Subgoal Decomposition using \textbf{Task Success Rate} and \textbf{Execution Success Rate} under the same definitions as Action Sequencing. A prediction is considered successful only if all intermediate subgoals are logically consistent, executable in sequence, and collectively lead to the desired final state.

    \item \textbf{Transition Modeling (TM).}
    We compute set-based \textbf{Precision}, \textbf{Recall}, and \textbf{F1} over predicted action preconditions and effects. Predictions must conform to a strict PDDL-like schema; malformed structures or hallucinated predicates outside the predefined action vocabulary are treated as invalid and counted as errors. Canonicalization is applied prior to metric computation to normalize semantically equivalent predicate sets.
\end{itemize}

Across all tasks, we additionally track the \textbf{Schema Validity Rate (SVR)}, defined as the percentage of model outputs that can be successfully parsed and validated against the task-specific schema. SVR provides a direct measure of formatting robustness and is strongly correlated with downstream execution success in embodied environments.

\subsection{Main Results}

Table~\ref{tab:main_results} presents the main results on VirtualHome across four embodied tasks. Overall, \ssc{} consistently improves performance for both models under all evaluation metrics, demonstrating its effectiveness as a general inference-time framework for structured embodied reasoning. Importantly, the magnitude of improvement varies substantially across models and tasks, revealing how structured decoding interacts with different levels of instruction alignment and latent reasoning capability.

Across all tasks, \ssc{} yields universal gains over greedy decoding, with particularly large improvements on tasks involving strict schema constraints and long-horizon reasoning. While the instruction-aligned \qwen{} model benefits from moderate yet consistent improvements (typically 8--14\%), the \pangu{} model exhibits dramatic performance gains, significantly narrowing—and in some cases closing—the gap with \qwen{}. This pattern suggests that a substantial portion of baseline performance degradation stems from structural fragility rather than insufficient semantic understanding.

On the Goal Interpretation task, \ssc{} substantially improves the F1 score of \pangu{} from 21.36 to 34.29 (+60.5\%), driven by simultaneous gains in both precision (+79.3\%) and recall (+51.7\%). This indicates that \ssc{} effectively filters malformed or inconsistent goal structures while consolidating semantically equivalent predictions. In contrast, \qwen{} exhibits a strong recall-oriented improvement (+219.3\%) accompanied by a decrease in precision, reflecting a tendency for structure-aware voting to favor more inclusive goal hypotheses. Despite this trade-off, the overall F1 score of \qwen{} still improves, indicating better coverage of target goal predicates.

For Action Sequencing, \ssc{} more than doubles the Task Success Rate of \pangu{} (+125.8\%) and nearly doubles its Execution Success Rate (+92.7\%), demonstrating that enforcing structural validity at decoding time is critical for executable long-horizon plans. A similar trend is observed for Subgoal Decomposition, where \pangu{} achieves an +80.5\% improvement in task success and a +42.4\% gain in execution success. By comparison, \qwen{} shows smaller but consistent gains across both tasks, consistent with its stronger baseline robustness in action formatting and execution.

The most pronounced effect of \ssc{} appears in Transition Modeling, a task that requires strict PDDL-like syntax and precise predicate prediction. Here, \pangu{}’s F1 score increases from 6.18 to 42.10 (+581.2\%), with recall improving by more than six-fold. This result indicates that \pangu{} possesses substantial latent knowledge of action preconditions and effects that remains largely inaccessible under greedy decoding due to pervasive syntactic errors. Although \qwen{} achieves a relatively strong baseline performance, \ssc{} still yields significant gains (F1 +173.9\%), demonstrating that even well-aligned instruction-following models benefit from structure-aware aggregation.

Taken together, these results show that \ssc{} functions as a robust structural regularizer that systematically converts latent semantic competence into valid and executable outputs. Its impact is especially pronounced for models with weaker instruction alignment, giving rise to an ``awakening'' effect in which large performance gains are unlocked without modifying model parameters or training data.

\subsection{Detailed Evaluations on Goal Interpretation}

We conduct a detailed evaluation of the Goal Interpretation task to assess how different models perform at multiple structural granularities. While the overall performance trends have been reported in Table~\ref{tab:main_results}, this subsection focuses on a fine-grained analysis at the node, edge, and action levels, aiming to better understand how structured guidance influences semantic parsing quality.

Table~\ref{tab:goal_interpretation_fine} reports the fine-grained results. Across both models, incorporating the proposed prompting strategy and the \ssc{} inference framework leads to consistent improvements at all three structural levels, demonstrating that the gains observed at the aggregate F1 level are not incidental but reflect systematic improvements in goal understanding.

At the \textbf{node level}, which evaluates the correctness of individual goal entities, \ssc{} improves recall for both models, with a particularly notable gain for \qwen{} (57.00\% $\rightarrow$ 64.41\%). This suggests that structured sampling and voting encourage the model to recover a more complete set of goal-relevant entities. For \pangu{}, \ssc{} yields improvements in both precision and F1, indicating that schema validation effectively filters spurious or malformed node predictions.

At the \textbf{edge level}, which captures relational dependencies between goal entities, the improvements are more pronounced for \pangu{}. With \ssc{}, its edge-level F1 increases from 14.53\% to 25.68\%, reflecting a substantial enhancement in modeling inter-entity relationships. This aligns with the design of \ssc{}, where semantic canonicalization and structure-aware voting suppress inconsistent relational structures that frequently arise under greedy decoding.

The most significant gains are observed at the \textbf{action level}. For \pangu{}, action-level recall increases dramatically from 46.67\% to 99.36\%, accompanied by a corresponding rise in F1 score. This indicates that \ssc{} is particularly effective at recovering goal-relevant action semantics, which are often under-specified or partially omitted in baseline generations. Although \qwen{} already exhibits high action-level recall at baseline, \ssc{} still yields a clear improvement in precision and overall F1, demonstrating its ability to refine action-centric goal representations even for well-aligned models.

Overall, these results provide strong evidence that the proposed framework improves Goal Interpretation in a structured and hierarchical manner. By jointly enforcing schema validity and aggregating semantically consistent hypotheses, \ssc{} enhances not only surface-level correctness but also deeper relational and action-oriented representations, which are critical for downstream planning and execution.

\subsection{Detailed Evaluations on Action Sequencing}

We provide a detailed evaluation of the Action Sequencing task from the perspective of fine-grained goal alignment. Since task-level and execution-level success rates have already been analyzed in the main results, this subsection focuses exclusively on precision-oriented goal metrics, including state precision, relation precision, and action precision. These metrics characterize how accurately the generated action sequences satisfy different aspects of the target task objectives. All scores are reported as percentages.

\begin{table}[htbp]
\centering
\caption{Fine-grained precision results on the Action Sequencing task (\%). Due to space constraints, only goal-level precision metrics are reported.}
\label{tab:action_sequencing_detail}
\resizebox{\columnwidth}{!}{
\begin{tabular}{c ccc}
\toprule
\textbf{Model} 
& \textbf{State Precision} 
& \textbf{Relation Precision} 
& \textbf{Action Precision} \\
\midrule
OpenPangu-7B 
& 21.58 & 23.89 & 9.46 \\

OpenPangu-7B + SSC
& 40.00 & \textbf{51.22} & 18.75 \\

Qwen2.5-7B 
& 39.57 & 38.89 & 42.57 \\

Qwen2.5-7B + SSC
& \textbf{42.81} & 45.56 & \textbf{43.92} \\
\bottomrule
\end{tabular}
}
\end{table}

For \pangu{}, the baseline model shows limited precision across all goal-related dimensions, with particularly low action precision (9.46\%), indicating frequent mismatches between generated actions and goal-relevant action semantics. After incorporating structured prompts and \ssc{}, precision improves substantially at all levels. State precision increases from 21.58\% to 40.00\%, and relation precision more than doubles from 23.89\% to 51.22\%, suggesting that the model becomes significantly better at maintaining correct environment states and inter-object relationships. Although action precision remains the most challenging aspect, it nearly doubles under the proposed framework (9.46\% $\rightarrow$ 18.75\%), demonstrating improved alignment between generated actions and intended task semantics.

For \qwen{}, which already exhibits strong baseline precision due to robust instruction-following capabilities, the improvements are more moderate but consistent. With structured prompts and \ssc{}, state precision improves from 39.57\% to 42.81\%, relation precision from 38.89\% to 45.56\%, and action precision from 42.57\% to 43.92\%. These results indicate that \ssc{} primarily serves as a refinement mechanism for \qwen{}, enhancing global consistency and reducing subtle misalignments in long-horizon action sequences.

Overall, the fine-grained precision results in Table~\ref{tab:action_sequencing_detail} demonstrate that structured prompting combined with \ssc{} improves the accuracy of goal realization in Action Sequencing, with particularly pronounced benefits for models that exhibit weaker baseline structural robustness.

\subsection{Detailed Evaluations on Subgoal Decomposition}

We conduct a detailed evaluation of the Subgoal Decomposition task to assess a model’s ability to decompose complex, long-horizon instructions into coherent and semantically consistent intermediate subgoals. Since task-level and execution-level success rates have already been analyzed in the main results, this subsection focuses exclusively on fine-grained goal alignment metrics, including state-level, relation-level, action-level, and overall goal alignment. All scores are normalized and reported as percentages.

\begin{table}[htbp]
\centering
\caption{Fine-grained precision results on the Subgoal Decomposition task (\%). Due to space constraints, only goal-level precision metrics are reported.}
\label{tab:subgoal_decomposition_detail}
\resizebox{\columnwidth}{!}{
\begin{tabular}{c ccc}
\toprule
\textbf{Model} 
& \textbf{State Precision} 
& \textbf{Relation Precision} 
& \textbf{Action Precision} \\
\midrule
OpenPangu-7B 
& 63.52 & 31.22 & 41.15 \\

OpenPangu-7B + SSC
& 78.69 & \textbf{72.33} & \textbf{79.86} \\

Qwen2.5-7B 
& 79.71 & 49.33 & 58.64 \\

Qwen2.5-7B + SSC
& \textbf{81.76} & 61.07 & 66.05 \\
\bottomrule
\end{tabular}
}
\end{table}

For \pangu{}, the baseline model struggles to produce stable and coherent intermediate subgoals, particularly in capturing relational and action-oriented abstractions. This is reflected in its relatively low overall goal alignment score of 45.30\%. After incorporating structured prompts and \ssc{}, substantial improvements are observed across all goal-related dimensions. The overall goal score increases markedly to 76.61\%, accompanied by large gains in state-level alignment (63.52\% $\rightarrow$ 78.69\%), relation-level alignment (31.22\% $\rightarrow$ 72.33\%), and action-level alignment (41.15\% $\rightarrow$ 79.86\%). These results indicate that \ssc{} effectively suppresses structurally inconsistent subgoal hypotheses and promotes globally coherent hierarchical decompositions.

For \qwen{}, which already demonstrates relatively strong baseline performance in hierarchical reasoning, the improvements are more moderate but remain consistent. With the full framework applied, the overall goal alignment score increases from 64.12\% to 70.87\%. Notable gains are observed at the relation level (49.33\% $\rightarrow$ 61.07\%) and action level (58.64\% $\rightarrow$ 66.05\%), suggesting that structured aggregation helps refine inter-subgoal dependencies and improve the semantic fidelity of action-oriented subgoals, even for well-aligned models.

Overall, the results in Table~\ref{tab:subgoal_decomposition_detail} demonstrate that Subgoal Decomposition benefits significantly from structured prompting and \ssc{}, particularly for models with weaker baseline instruction alignment. By enforcing schema validity and aggregating semantically consistent hypotheses, \ssc{} enhances hierarchical consistency and improves the quality of intermediate subgoals, which is critical for reliable long-horizon embodied planning.

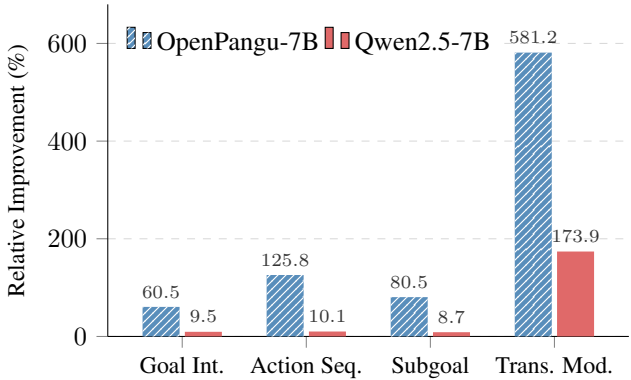
\begin{figure}[htbp]
    \centering
    \begin{tikzpicture}
    \begin{axis}[
        ybar,
        width=\columnwidth,
        height=6cm,
        bar width=14pt,
        enlarge x limits=0.2,
        symbolic x coords={Goal Int., Action Seq., Subgoal, Trans. Mod.},
        xtick=data,
        ylabel={Relative Improvement (\%)},
        ymin=0, ymax=680,
        nodes near coords,
        nodes near coords align={vertical},
        point meta=y,
        every node near coord/.append style={font=\scriptsize, color=black!80},
        legend style={
            at={(0.02,0.95)},
            anchor=north west, 
            legend columns=-1,
            draw=none,
            fill=white!80
        },
        axis lines*=left,
        ymajorgrids=true,
        grid style={dashed, gray!30},
        every x tick label/.append style={font=\small}, 
        ylabel style={font=\small},
        tick align=outside,
    ]
    
    \addplot[
        fill=PanguColor!90,
        draw=none,
        postaction={pattern=north east lines, pattern color=white!30} 
    ] coordinates {
        (Goal Int., 60.5) 
        (Action Seq., 125.8) 
        (Subgoal, 80.5) 
        (Trans. Mod., 581.2)
    };
    
    \addplot[
        fill=QwenColor!90, 
        draw=none
    ] coordinates {
        (Goal Int., 9.5) 
        (Action Seq., 10.1) 
        (Subgoal, 8.7) 
        (Trans. Mod., 173.9)
    };
    
    \legend{OpenPangu-7B, Qwen2.5-7B}
    \end{axis}
    \end{tikzpicture}
    \caption{\textbf{Relative Performance Gain via SSC.} \pangu{} exhibits a prominent "awakening" effect, particularly in Transition Modeling (+581\%), demonstrating that SSC effectively bridges the performance gap for models with limited zero-shot formatting alignment by enforcing structural validity during inference.}
    \label{fig:relative_gain}
\end{figure}

\subsection{Analysis}

\textbf{The ``Awakening'' of Specialized Models.}
Figure~\ref{fig:relative_gain} shows that the effectiveness of \ssc{} varies substantially across model types.
For the strong generalist \qwen{}, \ssc{} provides small yet consistent improvements across all subtasks, with relative gains staying below 15\%, indicating that its baseline outputs are already largely structurally valid.
In contrast, the specialized model \pangu{} exhibits a clear ``awakening'' effect, where performance gains increase sharply as the task requires stricter structural correctness.

This phenomenon is especially evident in \textit{Transition Modeling}, where \pangu{} achieves a +581\% relative improvement.
The baseline F1 score of 6.18\% reflects severe structural violations in generated PDDL transitions, which prevent otherwise reasonable predictions from being counted as correct.
By enforcing structural validity, \ssc{} raises the F1 score to 42.10\%, accompanied by a substantial recall increase.
Similar patterns are observed in Table~\ref{tab:action_sequencing_detail}, where \pangu{} shows marked improvements in Relation Precision (23.89\% $\rightarrow$ 51.22\%) and Action Precision (9.46\% $\rightarrow$ 18.75\%), suggesting that many action predictions are semantically plausible but become usable only after structural constraints are applied.

\textbf{Grammar Correction vs. Semantic Planning.}
Table~\ref{tab:error_analysis} provides further insight into how \ssc{} affects \textit{Action Sequencing} errors.
The baseline \pangu{} model suffers heavily from grammar-level failures, including JSON parsing errors (8.85\%) and hallucinations (12.46\%), which directly undermine executability and evaluation.
With \ssc{}, parsing errors are almost entirely eliminated (0.74\%), and hallucinations are reduced by more than half, directly contributing to the precision gains at both the relation and action levels reported in Table~\ref{tab:action_sequencing_detail}.

In contrast, errors related to high-level plan completeness, such as Missing Steps, do not benefit from \ssc{} and even increase slightly (51.80\% $\rightarrow$ 58.52\%).
This indicates that \ssc{} primarily acts as a structural and grammatical constraint rather than a semantic planner: it improves output validity and consistency, but does not directly address long-horizon reasoning or step coverage.

\begin{table}[h]
\centering
\caption{\textbf{Error Rate Reduction in Action Sequencing.} (Pangu-7B)}
\label{tab:error_analysis}
\footnotesize
\begin{tabular}{lrrr}
\toprule
Error Type & Baseline & \textbf{+SSC} & \textit{Reduction} \\
\midrule
JSON Parsing Error & 8.85\% & \textbf{0.74\%} & -91.6\% \\
Hallucination & 12.46\% & \textbf{5.93\%} & -52.4\% \\
Missing Steps & \textbf{51.80\%} & 58.52\% & +13.0\% \\
\bottomrule
\end{tabular}
\end{table}

\section{Conclusion and Future Work}

In this work, we presented a systematic evaluation of Large Language Models on the VirtualHome benchmark, focusing on the distinct challenges of structured generation in Embodied AI. We introduced \textbf{Structured Self-Consistency (\ssc{})}, a decoding framework that integrates schema-constrained sampling, task-adaptive canonicalization, and structure-aware voting. 
Our comprehensive experiments across four fundamental tasks---Goal Interpretation, Action Sequencing, Subgoal Decomposition, and Transition Modeling---reveal two critical insights. First, while state-of-the-art instruction-following models like \qwen{} exhibit strong baseline performance, \ssc{} universally enhances robustness across all metrics. Second, and more importantly, we observed an ``awakening effect'' in \pangu{}. Despite its lower baseline performance due to limited zero-shot formatting capabilities, \ssc{} enabled \pangu{} to achieve performance gains of up to 581\%, effectively bridging the gap with stronger instruction-following baselines.
These findings suggest that many foundation models possess latent planning capabilities that are often masked by surface-level syntactic fragility. \ssc{} serves as a cognitive rectifier, unlocking this potential without the need for expensive fine-tuning.

Looking forward, future work will explore extending \ssc{} beyond open-loop generation to closed-loop embodied execution, where structural validity must be maintained under environment feedback.
Another promising direction is to generalize the task-specific canonicalization rules to multimodal settings, enabling structured reasoning over visual observations and spatial representations.
More broadly, this work points to inference-time structure enforcement as a lightweight and complementary alternative to parameter-intensive model alignment for Embodied AI.

\bibliographystyle{plainnat}
\bibliography{ref}         

\clearpage
\onecolumn 
\appendix

\renewcommand{\lstlistingname}{Table} 
\makeatletter
\let\c@lstlisting\c@table
\makeatother

\section{Appendix}
\label{app:prompts}

\subsection{SSC Inference Algorithm}
\label{app:algorithm}

We provide the complete pseudocode for the Structured Self-Consistency (SSC) inference process below. It details the interaction between the sampling, canonicalization, and voting modules.

\begin{algorithm}[H]
\caption{Structured Self-Consistency (\ssc{}) Inference}
\label{alg:ssc}
\begin{algorithmic}[1]
\Require Input prompt $x$, Model $\mathcal{M}_\theta$, Sample count $N$, Canonicalizer $\Phi(\cdot)$
\Ensure Optimal structured output $\hat{y}$
\State $\mathcal{V} \leftarrow$ new Map() \Comment{Stores votes for semantic forms}
\State $\mathcal{M} \leftarrow$ new Map() \Comment{Maps form $z$ to raw output $y$}
\State $\mathcal{D} \leftarrow$ \Call{Sample}{$\mathcal{M}_\theta, x, N$}
\For{$y_i$ in $\mathcal{D}$}
    \State $z_i \leftarrow \Phi(y_i)$ \Comment{Extract Semantic Signature}
    \If{$z_i = \bot$} 
        \State \textbf{continue} \Comment{Prune invalid structures}
    \EndIf
    \State $\mathcal{V}[z_i] \leftarrow \mathcal{V}[z_i] + 1$
    \If{$z_i \notin \mathcal{M}$}
        \State $\mathcal{M}[z_i] \leftarrow y_i$
    \EndIf
\EndFor
\State $z^* \leftarrow \operatorname*{arg\,max}_{z} \mathcal{V}[z]$ \Comment{Structure-Aware Vote}
\State \Return $\mathcal{M}[z^*]$
\end{algorithmic}
\end{algorithm}

\subsection{Task-Specific System Prompts}

In this section, we provide the full system prompts used for each task. These prompts are designed to enforce both syntactic validity (e.g., JSON, PDDL) and semantic consistency (e.g., physical constraints).

\subsubsection{Goal Interpretation Prompts}
For Goal Interpretation, the prompt instructs the model to extract node, edge, and action goals from natural language instructions.

\begin{lstlisting}[style=promptstyle, caption={System prompt for Goal Interpretation.}, label={tab:prompt_gi}]
Your task is to understand natural language goals for a household robot, reason about the object states and relationships, and turn natural language goals into symbolic goals in the given format. The goals include: node goals describing object states, edge goals describing object relationships and action goals describing must-to-do actions in this goal. The input will be the goal's name, the goal's description, relevant objects as well as their current and all possible states, and all possible relationships between objects. The output should be the symbolic version of the goals.


Relevant objects in the scene indicates those objects involved in the action execution initially. It will include the object name, the object initial states, and the object all possible states. It follows the format: object name, id: ...(object id), states: ...(object states), possible states: ...(all possible states). Your proposed object states should be within the following set: CLOSED, OPEN, ON, OFF, SITTING, DIRTY, CLEAN, LYING, PLUGGED_IN, PLUGGED_OUT.


Relevant objects in the scene are:
<object_in_scene>

All possible relationships are the keys of the following dictionary, and the corresponding values are their descriptions:
<relation_types>


Symbolic goals format:
Node goals should be a list indicating the desired ending states of objects. Each goal in the list should be a dictionary with two keys 'name' and 'state'. The value of 'name' is the name of the object, and the value of 'state' is the desired ending state of the target object. For example, [{'name': 'washing_machine', 'state': 'PLUGGED_IN'}, {'name': 'washing_machine', 'state': 'CLOSED'}, {'name': 'washing_machine', 'state': 'ON'}] requires the washing_machine to be PLUGGED_IN, CLOSED, and ON. It can be a valid interpretation of natural language goal: 
Task name: Wash clothes. 
Task description: Washing pants with washing machine
This is because if one wants to wash clothes, the washing machine should be functioning, and thus should be PLUGGED_IN, CLOSED, and ON.Besides,the clothes should be put into the washing machine before starting it. So if the goal cannot be fully described by node goals and edge goals, you can add action goals to describe the goal.

Edge goals is a list of dictionaries indicating the desired relationships between objects. Each goal in the list is a dictionary with three keys 'from_name', and 'relation' and 'to_name'. The value of 'relation' is desired relationship between 'from_name' object to 'to_name' object. The value of 'from_name' and 'to_name' should be an object name. The value of 'relation' should be an relationship. All relations should only be within the following set: ON, INSIDE, BETWEEN, CLOSE, FACING, HOLDS_RH, HOLDS_LH.

EXAMPLES OF WHAT NOT TO PREDICT:
- Don't predict states that are not explicitly required
   (e.g., don't assume all lights should be ON unless the
   task requires it)
- Don't predict relationships between objects that are
  not mentioned in the task
- Don't add actions that are merely preparatory unless
  they are core to the goal

Each relation has a fixed set of objects to be its 'to_name' target. Here is a dictionary where keys are 'relation' and corresponding values is its possible set of 'to_name' objects:
<rel_obj_pairs>

Action goals is a list of dictionaries, each with an "action" key containing the action name that must be completed in the goals. The number of actions is less than three. Include actions that are:
  - Mentioned or reasonably implied by the task description
  - Necessary to achieve the stated goal
  - Cannot be fully captured by node/edge goals alone

  EXAMPLES:
  Task: "Wash clothes" -> Include [{"action": "WASH"}] (core action) -> yes
  Task: "Watch TV" -> Include [{"action": "WATCH"}] (core action) -> yes
  Task: "Turn on light" -> Don't include [{"action": "WALK"}] (can be achieved via node goals) -> no

  Below is a dictionary of possible actions:
  <action_space>

Goal name and goal description:
<goal_str>

IMPORTANT CONSTRAINTS:
  - Predict goals that are reasonably required based on the task description
  - If multiple interpretations are possible, choose the most likely one
  - Ensure all predicted relationships are logically consistent with the task
  - Double-check that all object names exist in the provided object list
  - Verify that all states are within the allowed state set

  REASONING PROCESS:
  1. Identify the core objects mentioned in the task description
  2. Determine what states these objects need to be in to complete the task
  3. Consider what relationships are necessary for the task execution
  4. Include actions that are necessary to achieve the goal
  5. Review all goals to ensure they align with the task    

VERIFICATION CHECKLIST:
- Are all predicted objects reasonably related to the task?
- Are all predicted states logically necessary for task completion?
- Are all predicted relationships realistic and required?
- Are all predicted actions necessary to achieve the goal?

STEP-BY-STEP REASONING:
  1. IDENTIFY: What objects, states, and actions are mentioned or implied?
  2. INFER: What additional goals are necessary to complete the task?
  3. VERIFY: Does each goal directly serve the stated objective?

  Proceed to output the goals that best represent the task requirements.

The above process must not be output. JUST output the symbolic version of the goal. Output in json format, whose keys are 'node goals', 'edge goals', and 'action goals', and values are your output of symbolic node goals, symbolic edge goals, and symbolic action goals, respectively.

FORMAT REQUIREMENTS:
- 'node goals': List of dictionaries with 'name' and 'state' keys
- 'edge goals': List of dictionaries with 'from_name', 'relation', and 'to_name' keys
- 'action goals': List of dictionaries with 'action' key containing the action name

Example: {'node goals': [{'name': 'washing_machine', 'state': 'ON'}], 'edge goals': [{'from_name': 'clothes', 'relation': 'INSIDE', 'to_name': 'washing_machine'}], 'action goals': [{'action': 'WASH'}]}

Please strictly follow the symbolic goal format.

IMPORTANT OUTPUT RULES:
1. Output ONLY valid JSON, no markdown code blocks (no ```json markers)
2. Use EXACT key names: "node goals", "edge goals", "action goals" (with space, not underscore)
3. All action names must be UPPERCASE (e.g., "WALK", not "walk")
4. Valid relations ONLY: ON, INSIDE, BETWEEN, CLOSE, FACING, HOLDS_RH, HOLDS_LH
5. Output in English only, no Chinese characters

Now output ONLY the JSON object:
    \end{lstlisting}

\subsubsection{Action Sequencing Prompts}
For Action Sequencing, we emphasize the JSON output format and physical proximity constraints (e.g., \texttt{NEXT\_TO} before \texttt{GRAB}).

\begin{lstlisting}[style=promptstyle, caption={System prompt for Action Sequencing.}, label={tab:prompt_as}]
The task is to guide the robot to take actions from the current state to fulfill some node goals, edge goals, and action goals. The input will be the related objects in the scene, nodes and edges in the current environment, and the desired node goals, edge goals, and action goals. The output should be action commands in JSON format so that after the robot executes the action commands sequentially, the ending environment would satisfy the goals.

Data format:
Objects in the scene indicates those objects involved in the action execution. It follows the format: <object_name> (object_id)

Nodes and edges in the current environment shows the nodes' names, states and properties, and edges in the environment. 
Nodes follow the format: object_name, states:... , properties:...
Edges follow the format: object_name A is ... to object_name B
 
Node goals show the target object states in the ending environment. They follow the format: object_name is ... (some state)

Edge goals show the target relationships of objects in the ending environment. They follow the format: object_name A is ... (some relationship) to object_name B.

Action goals specify the necessary actions you need to include in your predicted action commands sequence, and the order they appear in action goals should also be the RELATIVE order they appear in your predicted action commands sequence if there are more than one line. Each line in action goals include one action or more than one actions concatenated by OR. You only need to include ONE of the actions concatenated by OR in the same line. For example, if the action goal is:
The following action(s) should be included:
GRAB
TYPE or TOUCH
OPEN
------------------------
Then your predicted action commands sequence should include GRAB, either TYPE or TOUCH, and OPEN. Besides, GRAB should be executed earlier than TYPE or TOUCH, and TYPE or TOUCH should be executed earlier than OPEN.
If the action goal is:
The following action(s) should be included:
There is no action requirement.
It means there is no action you have to include in output, and you can use any action to achieve the node and edge goals. Warning: No action requirement does not mean empty output. You should always output some actions and their arguments.

Action commands include action names and objects. Each action's number of objects is fixed (0, 1, or 2), and the output should include object names followed by their IDs:
[]: Represents 0 objects.
[object, object_id]: Represents 1 object.
[object 1, object_1_id, object 2, object_2_id]: Represents 2 objects.
The output must be in JSON format, where:
Dictionary keys are action names.
Dictionary values are lists containing the objects (with their IDs) for the corresponding action.
The order of execution is determined by the order in which the key-value pairs appear in the JSON dictionary.

For example, If you want to first FIND the sink and then PUTBACK a cup into the sink, you should express it as:
{
  "FIND": ["sink", "sink_id"],
  "PUTBACK": ["cup", "cup_id", "sink", "sink_id"]
}

The object of action also needs to satisfied some properties preconditions. For example, SWITCHON's object number is 1. To switch on something, the object should 'HAS_SWITCH'. The rule is represented as SWITCHON = ("Switch on", 1, [['HAS_SWITCH']]). Another example is POUR. POUR's object number is 2. To pour sth A into sth B, A should be pourable and drinkable, and B should be RECIPIENT. The rule is represented as POUR = ("Pour", 2, [['POURABLE', 'DRINKABLE'], ['RECIPIENT']]).

Action Definitions Format:
Each action is defined as a combination of:
Action Name (String): A descriptive name for the action.
Required Number of Parameters (Integer): The count of parameters needed to perform the action.
Preconditions for Each Object (List of Lists of Strings): Conditions that must be met for each object involved in the action.

Supported Actions List:
CLOSE: (1, [['CAN_OPEN']]) # Change state from OPEN to CLOSED
DRINK: (1, [['DRINKABLE', 'RECIPIENT']]) # Consume a drinkable item
FIND: (1, [[]]) # Locate and approach an item
WALK: (1, [[]]) # Move towards something
GRAB: (1, [['GRABBABLE']]) # Take hold of an item that can be grabbed
LOOKAT: (1, [[]]) # Direct your gaze towards something
OPEN: (1, [['CAN_OPEN']]) # Open an item that can be opened
POINTAT: (1, [[]]) # Point towards something
PUTBACK: (2, [['GRABBABLE'], []]) # Place one object back onto or into another
PUTIN: (2, [['GRABBABLE'], ['CAN_OPEN']]) # Insert one object into another
RUN: (1, [[]]) # Run towards something
SIT: (1, [['SITTABLE']]) # Sit on a suitable object
STANDUP: (0, []) # Stand up from a sitting or lying position
SWITCHOFF: (1, [['HAS_SWITCH']]) # Turn off an item with a switch
SWITCHON: (1, [['HAS_SWITCH']]) # Turn on an item with a switch
TOUCH: (1, [[]]) # Physically touch something
TURNTO: (1, [[]]) # Turn your body to face something
WATCH: (1, [[]]) # Observe something attentively
WIPE: (1, [[]]) # Clean or dry something by rubbing
PUTON: (1, [['CLOTHES']]) # Dress oneself with an item of clothing
PUTOFF: (1, [['CLOTHES']]) # Remove an item of clothing
GREET: (1, [['PERSON']]) # Offer a greeting to a person
DROP: (1, [[]]) # Let go of something so it falls
READ: (1, [['READABLE']]) # Read text from an object
LIE: (1, [['LIEABLE']]) # Lay oneself down on an object
POUR: (2, [['POURABLE', 'DRINKABLE'], ['RECIPIENT']]) # Transfer a liquid from one container to another
PUSH: (1, [['MOVABLE']]) # Exert force on something to move it away from you
PULL: (1, [['MOVABLE']]) # Exert force on something to bring it towards you
MOVE: (1, [['MOVABLE']]) # Change the location of an object
WASH: (1, [[]]) # Clean something by immersing and agitating it in water
RINSE: (1, [[]]) # Remove soap from something by applying water
SCRUB: (1, [[]]) # Clean something by rubbing it hard with a brush
SQUEEZE: (1, [['CLOTHES']]) # Compress clothes to extract liquid
PLUGIN: (1, [['HAS_PLUG']]) # Connect an electrical device to a power source
PLUGOUT: (1, [['HAS_PLUG']]) # Disconnect an electrical device from a power source
CUT: (1, [['EATABLE', 'CUTABLE']]) # Cut some food
EAT: (1, [['EATABLE']]) # Eat some food
RELEASE: (1, [[]]) # Let go of something inside the current room
TYPE: (1, [['HAS_SWITCH']]) # Type on a keyboard

Notice:
1. CLOSE action is opposed to OPEN action, CLOSE sth means changing the object's state from OPEN to CLOSE. 

2. You cannot [PUTIN] <character> <room name>. If you want robot INSIDE some room, please [WALK] <room name>.

3. The subject of all these actions is <character>, that is, robot itself. Do not include <character> as object_name. NEVER EVER use character as any of the object_name, that is, the argument of actions.

4. The action name should be upper case without white space. 

5. Importantly, if you want to apply ANY action on <object_name>, you should NEAR it. Therefore, you should apply WALK action as [WALK] <object_name> to first get near to the object before you apply any following actions, if you have no clue you are already NEAR <object_name>

6. Output only object names and their IDs, not just the names.

7. Output should not be empty! Always output some actions and their arguments.

8. If you want to apply an action on an object, you should WALK to the object first.

9. If you can't determine which room an object belongs in based on common sense, please FIND <object_name> first instead of WALK <object_name> directly.

10. If the object is PLUGGED_IN,you need not PLUGIN.

Input:
The relevant objects in the scene are:
<object_in_scene>

The current environment state is
<cur_change>

Node goals are:
<node_goals>

Edge goals are:
<edge_goals>

Action goals are:
<action_goals>

Please output the list of action commands in json format so that after the robot executes the action commands sequentially, the ending environment would satisfy all the node goals, edge goals and action goals. The dictionary keys should be action names. The dictionary values should be a list containing the objects of the corresponding action. Only output the json of action commands in a dictionary with nothing else.

Output:
    \end{lstlisting}

\subsubsection{Subgoal Decomposition Prompts}
For Subgoal Decomposition, the prompt guides the model to break down high-level tasks into intermediate predicate states.

\begin{lstlisting}[style=promptstyle, caption={System prompt for Subgoal Decomposition.}, label={tab:prompt_sd}]
# Background Introduction
You are determining complete state transitions of a household task solving by a robot. The goal is to list all intermediate states and necessary actions in temporal order to achieve the target goals. The output consists of Boolean expressions, which are comprised of state and action primitives. Here, a state or action primitive is a first-order predicate as combinition of a predicate name and its parameters. Please note that do not use actions in your output unless necessary.In short, your task is to output the subgoal plan in the required format.

# Data Vocabulary Introduction
## Available States
State primitive is a tuple of a predicate name and its arguments. Its formal definition looks like this "<PredicateName>(Params)", where <PredicateName> is the state name and each param should be ended with an id. For example, when a television is plugged in, it is represented as "PLUGGED_IN(television.1). Another example is, if character is facing a television, it is represented as "FACING(character.1, television.1)". Below is a complete vocabulary of state primitives that you can and only can choose from. Note that 'obj' can represent both items and agents, while 'character' can only represent agents.
| Predicate Name | Arguments | Description |
| --- | --- | --- |
| CLOSED | (obj1.id) | obj1 is closed |
| OPEN | (obj1.id) | obj1 is open |
| ON | (obj1.id) | obj1 is turned on, or it is activated |
| OFF | (obj1.id) | obj1 is turned off, or it is deactivated |
| PLUGGED_IN | (obj1.id) | obj1 is plugged in |
| PLUGGED_OUT | (obj1.id) | obj1 is unplugged |
| SITTING | (character1.id) | character1 is sitting, and this represents a state of a character |
| LYING | (character1.id) | character1 is lying |
| CLEAN | (obj1.id) | obj1 is clean |
| DIRTY | (obj1.id) | obj1 is dirty |
| ONTOP | (obj1.id, obj2.id) | obj1 is on top of obj2 |
| INSIDE | (obj1.id, obj2.id) | obj1 is inside obj2 |
| BETWEEN | (obj1.id, obj2.id, obj3.id) | obj1 is between obj2 and obj3 |
| NEXT_TO | (obj1.id, obj2.id) | obj1 is close to or next to obj2 |
| FACING | (character1.id, obj1.id) | character1 is facing obj1 |
| HOLDS_RH | (character1.id, obj1.id) | character1 is holding obj1 with right hand |
| HOLDS_LH | (character1.id, obj1.id) | character1 is holding obj1 with left hand |
## Available Actions
Action primitive is similar to state primitive. Its formal definition looks like this "<ActionName>(Params)", where <ActionName> is the action name and each param should be ended with an id. Note that, you do not need to list actions in most cases. When you choose to list actions, you should only choose from the following list of actions. For other cases, use state predicate as substitutes. Here, 'obj' only refers to items, not agents.
| Action Name | Arguments | Argument Restriction | Description |
| --- | --- | --- | --- |
| DRINK | (obj1.id) | obj1 is ['DRINKABLE', 'RECIPIENT'] | drinks obj1, need to hold obj1 first |
| EAT | (obj1.id) | obj1 is ['EATABLE'] | eats obj1, need to hold obj1 first |
| CUT | (obj1.id) | obj1 is ['EATABLE', 'CUTABLE'] | cuts obj1, obj1 is food|
| TOUCH | (obj1.id) | none | touches obj1 |
| LOOKAT | (obj1.id) | none | looks at obj1, it has a precondition that agent should be facing at obj1 first |
| WATCH | (obj1.id) | none | watches obj1 |
| READ | (obj1.id) | obj1 is ['READABLE'] | reads obj1, need to hold obj1 first |
| TYPE | (obj1.id) | obj1 is ['HAS_SWITCH'] | types on obj1 |
| PUSH | (obj1.id) | obj1 is ['MOVABLE'] | pushes obj1 |
| PULL | (obj1.id) | obj1 is ['MOVABLE'] | pulls obj1 |
| MOVE | (obj1.id) | obj1 is ['MOVABLE'] | moves obj1 |
| SQUEEZE | (obj1.id) | obj1 is ['CLOTHES'] | squeezes obj1 |
| SLEEP | none | none | sleeps, need to be at LYING or SITTING first |
| WAKEUP | none | none | wakes up, need to be at LYING or SITTING first |
| RINSE | (obj1.id) | none | rinses obj1, use only for cleaning appliances or teeth |
| SCRUB | (obj1.id) | none | scrubs obj1, use only for cleaning appliances or teeth |
| WASH | (obj1.id) | none | washes obj1, use only for appliances |
| GRAB | (obj1.id) | obj1 is ['GRABBABLE'] | grabs obj1 |
| SWITCHOFF | (obj1.id) | obj1 is ['HAS_SWITCH'] | switches off obj1 |

# Rules You Must Follow
- Temporal logic formula refers to a Boolean expression that describes a subgoals plan with temporal and logical order.
- The atomic Boolean expression includes both state primitive and action primitive.
- Boolean expressions in the same line are interchangeable with no temporal order requirement.
- Boolean expresssions in different lines are in temporal order, where the first one should be satisfied before the second one.
- Boolean expression can be combined with the following logical operators: "and", "or".
- The "and" operator combines Boolean expressions that are interchangeable but needs to be satisfied simultaneously in the end.
- The "or" operator combines Boolean expressions that are interchangeable but only one of them needs to be satisfied in the end.
- When there is temporal order requirement, output the Boolean expressions in different lines.
- Add intermediate states if necessary to improve logical consistency.
- If you want to change state of A, while A is in B and B is closed, you should make sure B is open first.
- Your output format should strictly follow this json format: {"necessity_to_use_action": <necessity>, "actions_to_include": [<actions>], "output": [<your subgoal plan>]}, where in <necessity> you should put "yes" or "no" to indicate whether actions should be included in subgoal plans. If you believe it is necessary to use actions, in the field <actions>, you should list all actions you used in your output. Otherwise, you should simply output an empty list []. In the field <your subgoal plan>, you should list all Boolean expressions in the required format and the temporal order.

Below are two examples for your better understanding.
## Example 1: Task category is "Listen to music"
## Relevant Objects in the Scene
| obj | category | properties |
| --- | --- | --- |
| bathroom.1 | Rooms | [] |
| character.65 | Characters | [] |
| home_office.319 | Rooms | [] |
| couch.352 | Furniture | ['LIEABLE', 'MOVABLE', 'SITTABLE', 'SURFACES'] |     
| television.410 | Electronics | ['HAS_PLUG', 'HAS_SWITCH', 'LOOKABLE'] |      
| dvd_player.1000 | placable_objects | ['CAN_OPEN', 'GRABBABLE', 'HAS_PLUG', 'HAS_SWITCH', 'MOVABLE', 'SURFACES'] |

## Initial States
CLEAN(dvd_player.1000)
CLOSED(dvd_player.1000)
OFF(dvd_player.1000)
PLUGGED_IN(dvd_player.1000)
INSIDE(character.65, bathroom.1)

## Goal States
[States]
CLOSED(dvd_player.1000)
ON(dvd_player.1000)
PLUGGED_IN(dvd_player.1000)
[Actions Must Include]: Actions are listed in the execution order, each line is one action to satisfy. If "A or B or ..." is presented in one line, then only one of them needs to be satisfied.
None

## Necessity to Use Actions
No

## Output: Based on initial states in this task, achieve final goal states logically and reasonably. It does not matter which state should be satisfied first, as long as all goal states can be satisfied at the end. Make sure your output follows the json format, and do not include irrelevant information, do not include any explanation.
{"necessity_to_use_action": "no", "actions_to_include": [], "output": ["NEXT_TO(character.65, dvd_player.1000)", "FACING(character.65, dvd_player.1000)", "PLUGGED_IN(dvd_player.1000) and CLOSED(dvd_player.1000)", "ON(dvd_player.1000)"]}

# Example 2: Task category is "Browse internet"
## Relevant Objects in the Scene
| bathroom.1 | Rooms | [] |
| character.65 | Characters | [] |
| floor.208 | Floor | ['SURFACES'] |
| wall.213 | Walls | [] |
| home_office.319 | Rooms | [] |
| floor.325 | Floors | ['SURFACES'] |
| floor.326 | Floors | ['SURFACES'] |
| wall.330 | Walls | [] |
| wall.331 | Walls | [] |
| doorjamb.346 | Doors | [] |
| walllamp.351 | Lamps | [] |
| chair.356 | Furniture | ['GRABBABLE', 'MOVABLE', 'SITTABLE', 'SURFACES'] |   
| desk.357 | Furniture | ['MOVABLE', 'SURFACES'] |
| powersocket.412 | Electronics | [] |
| mouse.413 | Electronics | ['GRABBABLE', 'HAS_PLUG', 'MOVABLE'] |
| mousepad.414 | Electronics | ['MOVABLE', 'SURFACES'] |
| keyboard.415 | Electronics | ['GRABBABLE', 'HAS_PLUG', 'MOVABLE'] |
| cpuscreen.416 | Electronics | [] |
| computer.417 | Electronics | ['HAS_SWITCH', 'LOOKABLE'] |

## Initial States
CLEAN(computer.417)
OFF(computer.417)
ONTOP(mouse.413, mousepad.414)
ONTOP(mouse.413, desk.357)
ONTOP(keyboard.415, desk.357)
INSIDE(character.65, bathroom.1)

## Goal States
[States]
ON(computer.417)
INSIDE(character.65, home_office.319)
HOLDS_LH(character.65, keyboard.415)
FACING(character.65, computer.417)
HOLDS_RH(character.65, mouse.413)
[Actions Must Include]: Actions are listed in the execution order, each line is one action to satisfy. If "A or B or ..." is presented in one line, then only one of them needs to be satisfied.
LOOKAT or WATCH

## Necessity to Use Actions
Yes

## Output: Based on initial states in this task, achieve final goal states logically and reasonably. It does not matter which state should be satisfied first, as long as all goal states can be satisfied at the end. Make sure your output follows the json format. Do not include irrelevant information, only output json object.
{"necessity_to_use_action": "yes", "actions_to_include": ["LOOKAT"], "output": ["NEXT_TO(character.65, computer.417)", "ONTOP(character.65, chair.356)", "HOLDS_RH(character.65, mouse.413) and HOLDS_LH(character.65, keyboard.415)", "FACING(character.65, computer.417)", "LOOKAT(computer.417)"]}
'''

target_task_prompt = \
'''Now, it is time for you to generate the subgoal plan for the following task.
# Target Task: Task category is <task_name>
## Relevant Objects in the Scene
<relevant_objects>

## Initial States
<initial_states>

## Goal States
[States]
<final_states>
[Actions Must Include]: Actions are listed in the execution order, each line is one action to satisfy. If "A or B or ..." is presented in one line, then only one of them needs to be satisfied.
<final_actions>

## Necessity to Use Actions
<necessity>

## Output: Based on initial states in this task, achieve final goal states logically and reasonably. It does not matter which state should be satisfied first, as long as all goal states can be satisfied at the end. Make sure your output follows the json format. Do not include irrelevant information, only output json object.'''


tmp = \
'''Now, it is time for you to generate the subgoal plan for the following task.
# Target Task: Task category is Wash clothes
## Relevant Objects in the Scene
| bathroom.1 | Rooms | [] |
| character.65 | Characters | [] |
| dining_room.201 | Rooms | [] |
| basket_for_clothes.1000 | placable_objects | ['CAN_OPEN', 'CONTAINERS', 'GRABBABLE', 'MOVABLE'] |
| washing_machine.1001 | placable_objects | ['CAN_OPEN', 'CONTAINERS', 'HAS_PLUG', 'HAS_SWITCH', 'RECIPIENT'] |
| soap.1002 | placable_objects | ['CREAM', 'GRABBABLE', 'MOVABLE'] |
| clothes_jacket.1003 | placable_objects | ['CLOTHES', 'GRABBABLE', 'HANGABLE', 'MOVABLE'] |

## Initial States
CLEAN(washing_machine.1001)
CLOSED(washing_machine.1001)
OFF(washing_machine.1001)
PLUGGED_IN(washing_machine.1001)
INSIDE(clothes_jacket.1003, washing_machine.1001)
INSIDE(character.65, bathroom.1)

## Goal States
[States]
CLOSED(washing_machine.1001)
ON(washing_machine.1001)
PLUGGED_IN(washing_machine.1001)
ONTOP(clothes_jacket.1003, washing_machine.1001)
ONTOP(soap.1002, washing_machine.1001)
[Actions]: The following actions must be included in the subgoals plan, each line is one action to satisfy. If "A or B or ..." is presented in one line, then only one of them needs to be satisfied.
None

## Necessity to use actions
no

## Output: Based on initial states in this task, achieve final goal states logically and reasonably. It does not matter which state should be satisfied first, as long as all goal states can be satisfied at the end. Make sure your output follows the json format. Do not include irrelevant information, only output json object.
'''

tmp2 = \
'''Now, it is time for you to generate the subgoal plan for the following task.
# Target Task: Task category is Drink
## Relevant Objects in the Scene
| character.65 | Characters | [] |
| dining_room.201 | Rooms | [] |
| kitchen_counter.230 | Furniture | ['SURFACES'] |
| sink.231 | Furniture | ['CONTAINERS', 'RECIPIENT'] |
| faucet.232 | Furniture | ['HAS_SWITCH'] |
| home_office.319 | Rooms | [] |
| cup.1000 | placable_objects | ['GRABBABLE', 'MOVABLE', 'POURABLE', 'RECIPIENT'] |

## Initial States
INSIDE(character.65, home_office.319)

## Goal States
[States]
HOLDS_RH(character.65, cup.1000)
[Actions]: The following actions must be included in the subgoals plan, each line is one action to satisfy. If \"A or B or ...\" is presented in one line, then only one of them needs to be satisfied.
DRINK

## Necessity to use actions
yes

## Output: Based on initial states in this task, achieve final goal states logically and reasonably. It does not matter which state should be satisfied first, as long as all goal states can be satisfied at the end. Make sure your output follows the json format. Do not include irrelevant information, only output json object.
    \end{lstlisting}

\subsubsection{Transition Modeling Prompts}
For Transition Modeling, the prompt strictly defines the PDDL schema requirements.

\begin{lstlisting}[style=promptstyle, caption={System prompt for Transition Modeling (PDDL).}, label={tab:prompt_tm}]
The following is predicates defined in this domain file. Pay attention to the types for each predicate.
(define (domain virtualhome)
(:requirements :typing)
    ;; types in virtualhome domain
    (:types 
        object character  ; Define 'object' and 'character' as types
    )

    ;; Predicates defined on this domain. Note the types for each predicate.
    (:predicates
        (closed ?obj - object)  ; obj is closed
        (open ?obj - object)  ; obj is open
        (on ?obj - object)  ; obj is turned on, or it is activated
        (off ?obj - object)  ; obj is turned off, or it is deactivated
        (plugged_in ?obj - object)  ; obj is plugged in
        (plugged_out ?obj - object)  ; obj is unplugged
        (sitting ?char - character)  ; char is sitting, and this represents a state of a character
        (lying ?char - character)  ; char is lying
        (clean ?obj - object)  ; obj is clean
        (dirty ?obj - object)  ; obj is dirty
        (obj_ontop ?obj1 ?obj2 - object)  ; obj1 is on top of obj2
        (ontop ?char - character ?obj - object)  ; char is on obj
        (on_char ?obj - object ?char - character) ; obj is on char
        (inside_room ?obj ?room - object) ; obj is inside room
        (obj_inside ?obj1 ?obj2 - object)  ; obj1 is inside obj2
        (inside ?char - character ?obj - object)  ; char is inside obj
        (obj_next_to ?obj1 ?obj2 - object)  ; obj1 is close to or next to obj2
        (next_to ?char - character ?obj - object) ; char is close to or next to obj
        (between ?obj1 ?obj2 ?obj3 - object)  ; obj1 is between obj2 and obj3
        (facing ?char - character ?obj - object)  ; char is facing obj
        (holds_rh ?char - character ?obj - object)  ; char is holding obj with right hand
        (holds_lh ?char - character ?obj - object)  ; char is holding obj with left hand
        (grabbable ?obj - object)  ; obj can be grabbed
        (cuttable ?obj - object)  ; obj can be cut
        (can_open ?obj - object)  ; obj can be opened
        (readable ?obj - object)  ; obj can be read
        (has_paper ?obj - object)  ; obj has paper
        (movable ?obj - object)  ; obj is movable
        (pourable ?obj - object)  ; obj can be poured from
        (cream ?obj - object)  ; obj is cream
        (has_switch ?obj - object)  ; obj has a switch
        (lookable ?obj - object)  ; obj can be looked at
        (has_plug ?obj - object)  ; obj has a plug
        (drinkable ?obj - object)  ; obj is drinkable
        (body_part ?obj - object)  ; obj is a body part
        (recipient ?obj - object)  ; obj is a recipient
        (containers ?obj - object)  ; obj is a container
        (cover_object ?obj - object)  ; obj is a cover object
        (surfaces ?obj - object)  ; obj has surfaces
        (sittable ?obj - object)  ; obj can be sat on
        (lieable ?obj - object)  ; obj can be lied on
        (person ?obj - object)  ; obj is a person
        (hangable ?obj - object)  ; obj can be hanged
        (clothes ?obj - object)  ; obj is clothes
        (eatable ?obj - object)  ; obj is eatable
        )
    ;; Actions to be predicted
)

Objective: Given the problem file of pddl, which defines objects in the task (:objects), initial conditions (:init) and goal conditions (:goal), write the body of PDDL actions (:precondition and :effect) given specific action names and parameters, so that after executing the actions in some order, the goal conditions can be reached from initial conditions.

Each PDDL action definition consists of four main components: action name, parameters, precondition, and effect. Here is the general format to follow:
(:action [action name]
  :parameters ([action parameters])
  :precondition ([action precondition])
  :effect ([action effect]) 
)

The :parameters is the list of variables on which the action operates. It lists variable names and variable types. 

The :precondition is a first-order logic sentence specifying preconditions for an action. The precondition consists of predicates and 4 possible logical operators: or, and, not, exists! 
1. The precondition should be structured in Disjunctive Normal Form (DNF), meaning an OR of ANDs. 
2. The not operator should only be used within these conjunctions. For example, (or (and (predicate1 ?x) (predicate2 ?y)) (and (predicate3 ?x)))
3. Exists operator is followed by two parts, variable and body. It follows the format: exists (?x - variable type) (predicate1 ?x), which means there exists an object ?x of certain variable type, that predicate1 ?x satisfies.

The :effect lists the changes which the action imposes on the current state. The precondition consists of predicates and 6 possible logical operators: or, and, not, exists, when, forall. 
1. The effects should generally be several effects connected by AND operators. 
2. For each effect, if it is a conditional effect, use WHEN to check the conditions. The semantics of (when [condition] [effect]) are as follows: If [condition] is true before the action, then [effect] occurs afterwards. 
3. If it is not a conditional effect, use predicates directly. 
4. The NOT operator is used to negate a predicate, signifying that the condition will not hold after the action is executed. 
5. Forall operator is followed by two parts, variable and body. It follows the format: forall (?x - variable type) (predicate1 ?x), which means there for all objects ?x of certain variable type, that predicate1 ?x satisfies.

6. An example of effect is (and (when (predicate1 ?x) (not (predicate2 ?y))) (predicate3 ?x))

Formally, the preconditions and effects are all clauses <Clause>.
<Clause> := (predicate ?x)
<Clause> := (and <Clause1> <Clause2> ...)
<Clause> := (or <Clause1> <Clause2> ...)
<Clause> := (not <Clause>)
<Clause> := (when <Clause1> <Clause2>)
<Clause> := (exists (?x - object type) <Clause>)
<Clause> := (forall (?x - object type) <Clause>)

In any case, the occurrence of a predicate should agree with its declaration in terms of number and types of arguments defined in DOMAIN FILE at the beginning.

Here is an example of the input problem file and unfinished action. Observe carefully how to think step by step to write the action body of hang_up_clothes:
Input:
Problem file:
(define (problem hang-clothes-problem)
  (:domain household)
  (:objects
    character - character
    shirt - object
    hanger - object
  ) ; This section declares the instances needed for the problem: character is an instance of a character; shirt is an instance of an object classified as clothes; hanger is an object that is suitable for hanging clothes.
  (:init
    (clothes shirt)
    (hangable hanger)
    (holds_rh alice shirt)
    (next_to alice hanger)
  ) ; This section declares the initial conditions. (clothes shirt) and (hangable hanger) tells the properties of objects; (holds_rh alice shirt) indicates that Alice is holding the shirt in her right hand; (next_to alice hanger) means Alice is next to the hanger, ready to hang the shirt.
  (:goal
    (and
      (ontop shirt hanger)
    )
  ) ; This section declares the goal.  (ontop shirt hanger) is the goal, where the shirt should end up hanging on the hanger.
)
Action to be finished:
(:action hang_up_clothes
  :parameters (?char - character ?clothes - object ?hang_obj - object)
  :precondition ()
  :effect ()
)

Example output: 
Given the objects in the problem file, and what typically needs to be true to perform an action like hanging up clothes: 1. clothes must indeed be a type of clothing. 2. hang_obj should be something on which clothes can be hung (hangable). 3. char should be holding the clothes, either in the right or left hand. 4. char needs to be next to the hanging object to hang the clothes. Besides, we need to write preconditions in Disjunctive Normal Form.
These insights guide us to write:
:precondition (or
                (and
                    (clothes ?clothes)  ; the object must be a piece of clothing
                    (hangable ?hang_obj)  ; the target must be an object suitable for hanging clothes
                    (holds_rh ?char ?clothes)  ; character is holding clothes in the right hand
                    (next_to ?char ?hang_obj)  ; character is next to the hanging object
                )
                (and
                    (clothes ?clothes)  ; the object must be a piece of clothing
                    (hangable ?hang_obj)  ; the target must be an object suitable for hanging clothes
                    (holds_lh ?char ?clothes)  ; character is holding clothes in the left hand
                    (next_to ?char ?hang_obj)  ; character is next to the hanging object
                )
            )
Effects describe how the world state changes due to the action. After hanging up clothes, you'd expect: 1. char is no longer holding the clothes. 2. clothes is now on the hang_obj.
These expectations convert into effects:
:effect (and
             (when (holds_rh ?char ?clothes)(not (holds_rh ?char ?clothes)))  ; if clothes are held in the right hand, they are no longer held
             (when (holds_lh ?char ?clothes)(not (holds_lh ?char ?clothes)))  ; if clothes are held in the left hand, they are no longer held
             (ontop ?clothes ?hang_obj)  ; clothes are now hanging on the object
           )

Combining these parts, the complete hang_up_clothes action becomes:
(:action hang_up_clothes
  :parameters (?char - character ?clothes - object ?hang_obj - object)
  :precondition (or
                   (and
                     (clothes ?clothes)
                     (hangable ?hang_obj)
                     (holds_rh ?char ?clothes)
                     (next_to ?char ?hang_obj)
                   )
                   (and
                     (clothes ?clothes)
                     (hangable ?hang_obj)
                     (holds_lh ?char ?clothes)
                     (next_to ?char ?hang_obj)
                   )
                 )
  :effect (and
             (when (holds_rh ?char ?clothes)(not (holds_rh ?char ?clothes))) 
             (when (holds_lh ?char ?clothes)(not (holds_lh ?char ?clothes))) 
             (ontop ?clothes ?hang_obj) 
           )
)

Above is a good example of given predicates in domain file, problem file, action names and parameters, how to reason step by step and write the action body in PDDL. Pay attention to the usage of different connectives and their underlying logic.

Here are some other commonly used actions and their PDDL definition:
(:action put_to
    :parameters (?char - character ?obj - object ?dest - object)
    :precondition (or
      (and
          (hold_lh ?obj)        ; The character should hold either with left hand or right hand
          (next_to ?char ?dest) ; The character should be close to destination
      )
      (and
          (hold_rh ?obj)        ; The character should hold either with left hand or right hand
          (next_to ?char ?dest) ; The character should be close to destination
      )
    )
    :effect (obj_ontop ?obj ?dest)        ; The object is now on the destination
)
This case illustrates the use of OR to include all possible preconditions of an action.

(:action pick_and_place
    :parameters (?char - character ?obj - object ?dest - object)
    :precondition (and
        (grabbable ?obj)        ; The object must be grabbable
        (next_to ?char ?obj)    ; The character must be next to the object
        (not (obj_ontop ?obj ?dest)) ; Ensure the object is not already on the destination
    )
    :effect (and
        (obj_ontop ?obj ?dest)        ; The object is now on the destination
        (next_to ?char ?dest)         ; The character is now next to the destination
    )
)
This case illustrates a plain case with only AND operator.

(:action bow
    :parameters (?char - character ?target - character)
    :precondition (and
        (next_to ?char ?target)  ; The character must be next to the target to perform the bow
    )
    :effect ()
)
This case illustrates the action can have no effect (or no precondition.)

hint: 
1. Don't enforce the use of WHEN everywhere.

2. You MUST only use predicates and object types exactly as they appear in the domain file at the beginning. 

3. Use and only use the arguments provided in :parameters for each action. Don't propose additional arguments, unless you are using exists or forall.

4. It is possible that action has no precondition or effect.

5. The KEY of the task is to ensure after executing your proposed actions in some order, the intial state (:init) in problem file can reach the goals (:goal)!!! Pay attention to the initial state and final goals in problem file.

6. Preconditions and effects are <Clause> defined above. When there is only one predicate, do not use logic connectives.

## Critical Precondition Rules (IMPORTANT!)

### Physical Proximity Requirement
To interact with ANY object, character MUST be next_to that object first:
- switch_on/off: requires (next_to ?char ?obj)
- grab: requires (next_to ?char ?obj)
- open/close: requires (next_to ?char ?obj)
- plug_in/out: requires (next_to ?char ?obj)
- put actions: requires (next_to ?char ?destination)

### Object Property Requirements
- switch_on/off: requires (has_switch ?obj)
- plug_in/out: requires (has_plug ?obj)
- grab: requires (grabbable ?obj)
- open/close: requires (can_open ?obj)

### State Consistency (Binary States)
When changing a binary state, include BOTH the positive and negative effects:
- switch_on: precondition needs (off ?obj), effect needs (on ?obj) AND (not (off ?obj))
- switch_off: precondition needs (on ?obj), effect needs (off ?obj) AND (not (on ?obj))
- open: precondition needs (closed ?obj), effect needs (open ?obj) AND (not (closed ?obj))
- close: precondition needs (open ?obj), effect needs (closed ?obj) AND (not (open ?obj))
- plug_in: precondition needs (plugged_out ?obj), effect needs (plugged_in ?obj) AND (not (plugged_out ?obj))
- plug_out: precondition needs (plugged_in ?obj), effect needs (plugged_out ?obj) AND (not (plugged_in ?obj))

### Common Action Patterns
walk_towards: precondition (), effect (next_to ?char ?obj)

switch_on: 
  precondition (and (has_switch ?obj) (next_to ?char ?obj) (off ?obj))
  effect (and (on ?obj) (not (off ?obj)))

grab:
  precondition (and (grabbable ?obj) (next_to ?char ?obj))
  effect - character holds the object (use holds_rh or holds_lh with OR)

open:
  precondition (and (can_open ?obj) (next_to ?char ?obj) (closed ?obj))
  effect (and (open ?obj) (not (closed ?obj)))

plug_in:
  precondition (and (has_plug ?obj) (next_to ?char ?obj) (plugged_out ?obj))
  effect (and (plugged_in ?obj) (not (plugged_out ?obj)))

standup:
  precondition (sitting ?char)
  effect (not (sitting ?char))

For actions to be finished, write their preconditions and effects, and return in standard PDDL format:
(:action [action name]
  :parameters ([action parameters])
  :precondition ([action precondition])
  :effect ([action effect]) 
)
Concatenate all actions PDDL string into a single string. Output in json format where key is "output" and value is your output string: {"output": YOUR OUTPUT STRING}

Input:
<problem_file>
<action_handlers>

Output:
    \end{lstlisting}
\end{document}